\documentclass{article}
\usepackage{spconf,amsmath,graphicx}
\usepackage{amsmath,amssymb,amsfonts}
\usepackage{algorithmic}

\newcommand{\algoname}[0]{MTRGL}
\usepackage{algorithm}
\usepackage{subfigure}
\usepackage{multirow}

\title{{\algoname}: Effective Temporal Correlation Discerning through Multi-modal Temporal Relational Graph Learning  }

\name{$\text{Junwei Su}^2$, $\text{Shan Wu}^{1,*}$, $\text{Jinhui Li}^3$}
\address{$^1$ Hefei University of Technology, $^2$ The University of Hong Kong, $^3$ The University of Toronto }
\begin{document}

\maketitle

\ninept

\begin{abstract}
In this study, we explore the synergy of deep learning and financial market applications, focusing on pair trading. This market-neutral strategy is integral to quantitative finance and is apt for advanced deep-learning techniques. A pivotal challenge in pair trading is discerning temporal correlations among entities, necessitating the integration of diverse data modalities. Addressing this, we introduce a novel framework, Multi-modal Temporal Relation Graph Learning ({\algoname}). {\algoname} combines time series data and discrete features into a temporal graph and employs a memory-based temporal graph neural network. This approach reframes temporal correlation identification as a temporal graph link prediction task, which has shown empirical success. Our experiments on real-world datasets confirm the superior performance of {\algoname}, emphasizing its promise in refining automated pair trading strategies. \let\thefootnote\relax\footnotetext{*Corresponding author: Shan Wu, wus@hfut.edu.cn}\footnotetext{To appear as a conference paper at ICASSP 2024}
\end{abstract}

    
\section{Introduction}\label{sec:introduction}

\begin{figure}[t!]
    \centering
    \vspace{-4mm}
    \includegraphics[width=0.38\textwidth]{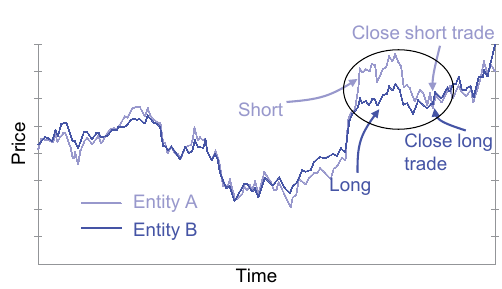}
        \vspace{-2mm}
    \caption{Illustration of the Pair Trading Strategy. Entities A and B are two correlated entities whose price tends to move together. The circled section indicates a temporal divergence of the price of the two entities due to market fluctuation or inefficiency. During the divergent period, the trader can simultaneously take a long position on entity B and a short position on entity A, and benefit from the divergence period when the movement of A and B converge again.}
    \label{fig:pair_trade}
    \vspace{-3mm}
\end{figure}

 Pair trading, a pivotal investment strategy, capitalizes on price disparities between related assets or markets \cite{gatev2006pairs}. Traders identify temporary price variances, executing simultaneous long and short positions on correlated assets. In a long position, a trader anticipates a price rise, while in a short position, they expect a decline, aiming to repurchase the asset at a reduced price later. This strategy, illustrated in Figure \ref{fig:pair_trade}, leverages market inefficiencies, enabling traders to profit from transient price deviations. By doing so, pair trading not only enhances market efficiency but also fosters improved returns and risk management in portfolios.

The essence of pair trading lies in identifying temporal correlations among assets or markets. Recognizing pairs with synchronized price movements, which tend to converge or diverge, is complex due to the market's dynamic nature and the vast number of potential pairs. This complexity demands robust quantitative analysis and advanced statistical techniques, further complicated by the evolving nature of asset relationships influenced by market shifts, regulations, and macroeconomic events.

Historically, experts manually identified these correlations. While traditional methods focused on statistical inferences, they often captured only basic data trends. The rise of machine learning has ushered in a new era, demonstrating its prowess in handling vast datasets, recognizing non-linear relationships, and outperforming traditional methods. However, the potential of machine learning in pair trading remains underexplored, necessitating further research.

Applying machine learning to pair trading introduces challenges. Firstly, while abundant data exists in finance, its simplicity can limit machine learning optimization. Advanced models thrive on complex, multi-dimensional inputs, surpassing traditional methods. Thus, integrating diverse data modalities is crucial. Secondly, the ever-changing financial markets, influenced by various factors, can drastically alter asset relationships. This demands models that can adapt to these temporal correlations, requiring advanced algorithms.

In this study, we examine the role of machine learning in the stock market by focusing on pair trading, introducing a novel framework, the multi-modal temporal relation graph learning ({\algoname}). The contributions of this paper are highlighted as follows:

$\bullet$  We probe the challenges associated with the application of machine learning in finance. We highlight the pivotal importance of integrating information from diverse modalities.

$\bullet$ The introduction of {\algoname}, a framework that seamlessly integrates various modalities. It features two key components: a mechanism for constructing a dynamic graph encapsulating both time series data (e.g., price trends) and discrete feature information (e.g., sector classifications), and a neural model powered by a memory-based dynamic graph neural network, an efficacious tool in temporal graph learning~\cite{dgb_neurips_D&B_2022}.

$\bullet$ Empirical analysis on real-world data, showcasing {\algoname}'s superiority in identifying temporal correlations. An ablation study further validates its effectiveness, offering insights for future machine learning applications in finance.

\section{RELATED WORK}

The rise in computational capabilities has enabled neural-network-based machine learning models to derive insights from complex datasets, including images~\cite{chui2022transfer,bui2022combination,waldchen2018machine}, languages~\cite{wan2021evaluation,khan2016survey,ofer2021language}, and networks~\cite{lee2022privacy,xia2021graph,gao2023runtime,su2023towards}. Financial markets, with their intricate dynamics, have become a focal point for machine learning applications. The allure of financial gains has driven increased research interest in this domain, with a comprehensive review available in~\cite{ozbayoglu2020deep}. Yet, much of the existing work has been confined to stock price data, a relatively straightforward time series, restricting the full exploitation of machine learning's capabilities.

Recent efforts~\cite{lopez2023can,qureshi2023chatgpt} have explored the capabilities of large language models like ChatGPT \cite{chen2023chatgpt,lund2023chatgpt} to extract features from financial news. However, the combination of these features with stock price data often remains rudimentary. In this study, we address this integration challenge, emphasizing temporal correlations for pair trading with {\algoname}. Our approach presents a refined method to integrate multi-modal information, creating structure within financial markets.

\begin{figure*}[t!]
    \centering
    \vspace{-8mm}
    \subfigure[Temporal Graph Construction Process of {\algoname}]{ \includegraphics[width=0.48\textwidth]{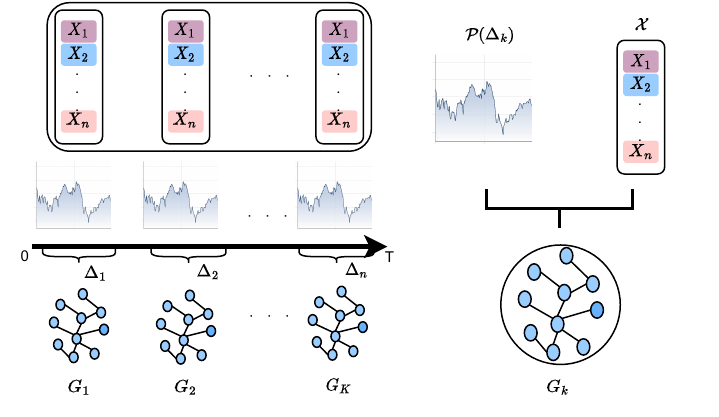}
    \label{fig:graph_construct}} 
    \subfigure[Training Pipeline of {\algoname}]{ \includegraphics[width=0.48\textwidth]{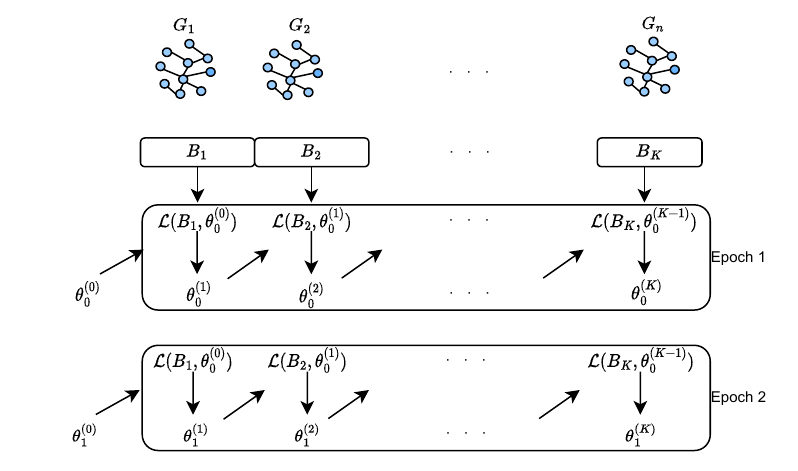}
    \label{fig:train_pipe}}    
    \vspace{-2mm}
    \caption{ Figure~\ref{fig:graph_construct} displays how the time series data spanning from $[0,T]$ is partitioned into a sequence of intervals $\{ \Delta_1,...,\Delta_n \}$. The latter portion of this figure presents how the information from these time series segments is amalgamated with feature data to create the sequence of temporal investment graphs $
    \{G_1,...,G_n\}$. On the other hand, Figure~\ref{fig:train_pipe} provides a visualization of {\algoname}'s training process. The constructed graphs are processed as event batches and employed for model updates. The loss of event batches $B_i$, denoted as $\mathcal{L}(B_i, \theta_{k}^{(j)})$, is calculated using the parameter $\theta_{k}^{(j)}$ obtained from the $k$-th epoch and $j$-th iteration. The parameters derived from the final iteration of each epoch serve as initial values for the subsequent epoch, denoted as $\theta_{k}^{(K-1)} = \theta_{k+1}^{(0)}$. } 
    \vspace{-2mm}
\end{figure*}

\section{Methodology}
Consider $\mathcal{A} = \{A_1, A_2, ..., A_n\}$ as a collection of $n$ entities under examination, where entities may represent companies, assets, or markets. Each entity $A_i$ is linked with a dynamic feature vector $X_i(t) = (x^{(1)}(t), x^{(2)}(t), ..., x^{(m)}(t))$, with each entry $x^{(j)}(t)$ signifying the $j$-th feature of entity $A_i$ at a given time $t$ (for instance, sector name, trading volume, market share, and so on). We use $\mathcal{X}(t)$ to denote the collection of feature vectors associated with $\mathcal{A}$ at time $t$. Moreover, $P_i(t): t \mapsto \mathbb{R}$ is the trading price time series of entity $A_i$, and we denote $P_i([0,t])$ as the time series information of entity $A_i$ in the time interval $[0,t]$ for a certain $t >0$. Let $\mathcal{P} = \{P_i(t)\}$ denote the set of time series linked to the entity set $\mathcal{A}$. Similarly, $\mathcal{P}([0,t]) = \{P_i([0,t])| P_i(.) \in \mathcal{P} \}$ represents the set of all entities' time series associated with the time interval $[0,t]$.

As previously discussed, a crucial aspect of pair trading is to discern the temporal correlation among entities based on their historical and feature information. More formally, if we denote $S(P_i(t),P_j(t))$, e.g., the normalized historical difference (NHD) \cite{gatev2006pairs}, as a predetermined correlation measure of two time series and $\gamma$ as a defined threshold, the aim is, given a set of entities $\mathcal{A}$, their feature vectors $\mathcal{X}$, and their time series information $\mathcal{P}([0,T])$, to identify pairs of entities $A_i, A_j$ such that $S(p_i([T,T+\delta]),p_j([T,T+\delta])) \geq \gamma$ for some $\delta > 0$.

\subsection{Temporal Graph Construction}
Given data from the time interval $[0,T]$, we initially divide this interval into $K$ smaller intervals, $\Omega = \{\Delta_i\}_{i=1,...,K}$. We assume that each time interval $\Delta_i$ has the same size of $|\Delta_i| = \delta = \frac{T}{K}$, i.e., $\Omega = \{[0,\delta),[\delta,2\delta),...,[(K-1)\delta,K \delta)\}$ for simplification.

Assuming $S( P_i(t), P_j(t)) \mapsto [0,1]$ is a given measure function (for simplicity, we use the NHD) for the correlation of two time series, and $\gamma$ is a given threshold, we construct the $k-th$ temporal graph $G_k$ associated with the time interval $\Delta_k$ as follows.

\begin{itemize}
\item We add a vertex $v_i$ to the graph for each entity $A_i$.
\item An edge $e_{ij}(t)$ is added between $v_i$ and $v_j$ only if \\$S( P_i(\Delta_k), P_j(\Delta_k)) \geq \gamma$.
\item We assign $t = (k-\frac{1}{2})\delta$ to the edge above.
\end{itemize}

The overall procedure of constructing the temporal graph is summarized in Algorithm~\ref{algo:graph_construction}. We denote $\mathcal{G} = \{G_i\}_{i=1,...,K}$ as the temporal graph sequence given by the above construction process. The set of vertices (entities) in $\mathcal{G}$ is denoted as $\mathcal{V} = \{1,...,n\}$. Additionally, following the practice in dynamic graph learning, we refer to an edge in $\mathcal{G}$ as an event.

\subsection{Temporal Graph Neural Networks}
Temporal graph neural networks (TGNNs) have shown substantial potency as neural models, particularly for forecasting temporal graphs~\cite{dgb_neurips_D&B_2022,rossi2021tgn,kumar2019jodie}. Memory-based TGNNs (MTGNNs), a subset of TGNNs, offer superior outcomes over memory-less alternatives~\cite{dgb_neurips_D&B_2022}. One salient feature of MTGNNs is the integration of a memory module, which serves as a filter, iteratively processing and refining data from new and historical graph events. As a result, MTGNNs capture extensive dependencies efficiently and deliver leading performance on a wide variety of dynamic graph-related tasks~\cite{zhang2023tiger}.

In our study, we employ and adapt MTGNNs as our neural model. Adhering to the settings in \cite{zhang2023tiger}\cite{rossi2020temporal}, we structure our MTGNN with an encoder-decoder configuration. The encoder of MTGNN comprises three primary modules: {\bf msg} (message), {\bf mem} (memory), and {\bf emb} (embedding) modules. The output from the embedding module is fed into a decoder (in this case, a simple two-layer MLP) to execute the inference task. The details of each module in the MTGNN are presented in the following sections.

\subsubsection{Encoder}
The encoder of MTGNN houses three modules (message, memory, and embedding), introduced separately for clarity. 

{\bf Message.} For each event (edge) that involves node $i$, a message is computed to refresh $i$’s memory. Given an interaction event $e_{ij}(t)$ between source node $i$ and target node $j$ at time $t$, two messages can be formulated:
\begin{equation}
\begin{split}
m_i(t) = \mathrm{msg}(s_i(t^-),s_j(t^-), e_{ij}(t), \psi(t-t_i')), \\
m_j(t) = \mathrm{msg}(s_j(t^-),s_i(t^-), e_{ij}(t), \psi(t-t_j')).
\end{split}
\end{equation}
Here, $s_i(t^-)$ and $s_j(t^-)$ are the memory states of node $i$ and $j$ just before the event at time $t$, $\mathrm{msg}(.)$ is the message function, and $t'_j$ is the timestamp of the last event involving node $j$. The time encoding function $\psi(.)$ \cite{xu2020inductive} maps the time interval into a $d$-dimension vector. For simplification, we use the popular identity message function that returns the concatenation of the input vectors \cite{zhang2023tiger}\cite{rossi2020temporal}.

{\bf Memory.} The memory of a node is updated for each event involving that node:
\begin{equation}
\begin{split}
s_i(t) = \mathrm{mem} (m_i(t), s_i (t^{-} )).
\end{split}
\end{equation}
In the case of interaction events that involve two nodes $i$ and $j$, the memories of both nodes are updated post the event. Here, $\mathrm{mem}$ is a learnable memory update function. In practice, we use the gated recurrent unit (GRU) \cite{cho2014learning}. The memory module functions as a filtering mechanism, learning and distilling information from both new and historical temporal graph data iteratively. Thus, it effectively captures long-range dependencies.

\begin{algorithm}[h]
  \caption{Training Procedure of {\algoname}}\label{algo:standard_train}
  \label{alg:train}
\begin{algorithmic}
    \STATE {\bfseries Input:} $\mathcal{G} = \{G_1,...,G_k\}$ \hfill // \COMMENT{The temporal graph sequence} \\
    \STATE {\bfseries Input:} $T$ \hfill // \COMMENT{Number of epoches} \\
   \STATE {\bfseries Initialization:} $S_{0} \leftarrow 0$ \COMMENT{Initialize memory vectors to be zero} \\
   \STATE {\bfseries Initialization:} $\theta_{0}^{(0)} \leftarrow 0$ \COMMENT{Initialize model parameter} \\
  \FOR{t=1 \textbf{to} T}
     \FOR{$G_i \in G_2,...,G_K$}
      \item $B_i^{+} \leftarrow \mathcal{E}_i$, \quad $B_i^{-} \leftarrow \text{Sample negative events}$ 
      \item   $\bar{B}_{i} = B_i^{-} \bigcup B_{i}^+ $   \COMMENT{Use the temporal batch from last iteration to update the memory and embedding}
      \item   $\bar{B}_{i-1} \leftarrow \text{Temporal batch from last iteration}$
      \item   $M_i = \mathrm{msg}(S_{i-1},\bar{B}_{i-1})$ \COMMENT{Compute messages for events} \\
      \item   $S_i = \mathrm{mem}(S_{i-1},M_i)$ \COMMENT{Update memory the message} \\
      \item   $H_i = \mathrm{emb}(S_i,A_i),$ \COMMENT{Compute the embedding} \\
      \item   Compute the loss $\mathcal{L}(H_i, B_{i}) $(given in Eq.~\eqref{eq:loss}) and update the model parameter with training algorithm (e.g., backpropagation and Adam) 
     \ENDFOR
  \ENDFOR
\end{algorithmic}
\end{algorithm} 

{\bf Embedding.} The aim of the temporal embedding module is to generate pre-jump representations $z(t^-)$ at any time $t$ before the arrival of the next event. We specifically employ an $L$-layer temporal graph attention network to gather neighbourhood information.

We first enhance the memory vector of entity $A_i$ with the node feature $z_i^{(0)} (t) = s_i (t) + X_i (t)$. This allows the model to utilize both the current memory $s_i(t)$ and the temporal node features $X_i(t)$. Then for each layer $1 \leq l \leq L$, we employ the multi-head attention~\cite{xu2020inductive} to aggregate neighbourhood information as follows:
\begin{equation}
\begin{split}
z_i^{(l)} &= \mathrm{mlp}^{(l)}(z^{(l-1)}||\tilde{z}_k^{(l)}),\\
\tilde{z}_k^{(l)} &= \mathrm{mha}^{(l)}(q^{(l)}_i (t),K^{(l)}_i(t), V^{(l)}_i (t)),\\
q^{(l)}(t) &= z_{i}^{(l-1)} ||\psi(0), \\
K_i^{(l)}(t) = V^{(l)}_i(t) &= \begin{bmatrix}
z_i^{(l-1)} || e_{\pi_i(1)}(t_{\pi_i(1)})|| \psi(t - t{\pi_i(1)})\\
...\\
z_i^{(l-1)} || e_{\pi_i(N)}(t_{\pi_i(N)})|| \psi(t - t_{\pi_i(N)})\\
\end{bmatrix}.
\end{split}
\end{equation}
Here, $||$ denotes vector concatenation, $\mathrm{mlp}(.)$ are one-layer MLP with hidden/output dimensions of $d$, and $\mathrm{mha}(.)$ are multi-head attention functions with queries $q(.)$, keys $K(.)$ and values $V(.)$. We limit the receptive fields to the most recent $N$ events for each node $i$, represented by $\pi_i = \{e_{\pi_i(1)}(t_{\pi_i(1)}), e_{\pi_i(1)}(t_{\pi_i(2)}),...,e_{\pi_i(N)}(t_{\pi_i(N)})\}$. In this context, $\pi$ is a permutation, and $\pi_{i}(.)$ are temporal neighbours of node $i$. The subscript $i$ in $\pi_{i}(.)$ is omitted, referring to an event from $i$ to $\pi_{i}(.)$ or vice versa.

\subsubsection{Decoder}
In our study, we simplify the problem of discerning temporal correlation into the temporal link prediction task. Given $z_i(t^-)$ and $z_j(t^-)$ as produced by the encoder, the decoder computes the probability of the event $e_{ij}(t)$ utilizing a two-layer MLP followed by the sigmoid function $\sigma(.)$,
\begin{equation}
\hat{p}_{ij}(t) = \sigma (\mathrm{MLP}(z_i(t^-)||z_j(t^-))).
\end{equation}

\subsection{Model Training and Inference}

The training process for the model leverages binary cross-entropy as the loss function and contrastive learning:
\begin{equation}\label{eq:loss}
\mathcal{L} = - \sum_{e_{ij}(t) \in \mathcal{E}}[\log \hat{p}_{ij}(t) + \log(1-\hat{p}_{ik}(t))],
\end{equation}
where a negative destination node is randomly sampled as $k$, and $\hat{p}_{ik}(t)$ is calculated in a similar manner using $z_i(t^{-})$ and $z_k (t^{-})$. This facilitates the neural model in learning contrastive signals derived from the generated graphs.

Moreover, we implement a temporal batch training method for the sake of efficiency. Successive events are segmented into a temporal batch, within which events are processed concurrently. For nodes associated with multiple events within a batch, we resort to the simplest recent message aggregator that retains only the most recent message for each node in the batch, following the practice from~\cite{zhang2023tiger}. In order to avoid information leakage (preventing the use of a batch's information to predict its own events), we employ a lag-one scheme in which temporal batch $B_{i-1}$ is utilized to refresh the memory state and create embeddings for predicting $B_{i}$. The comprehensive training procedure is provided in Algorithm~\ref{alg:train} and Figure~\ref{fig:train_pipe} gives a graphical illustration.

\vspace{-2mm}
\begin{algorithm}[h]
  \caption{Multi-modal Temporal Relation Graph Construction}\label{algo:graph_construction}
  \label{alg:train}
\begin{algorithmic}
   \STATE {\bfseries Input:} $\mathcal{A} = \{A_1,...,A_n\}$ \hfill // \COMMENT{ set of entities} \\
   \STATE {\bfseries Input:} $K$ \hfill // \COMMENT{Number of the time interval to be constructed} \\
   \STATE {\bfseries Input:} $\gamma$ \hfill // \COMMENT{The threshold for correlation } \\
   \STATE {\bfseries Input:} $S(.,.)$ \hfill // \COMMENT{a given covariance measure } \\
   \STATE {\bfseries Initialization:} $\mathcal{G} \leftarrow [.]$ \COMMENT{Initialize an empty list for storing graph} \\
  \STATE Divide $\mathcal{A}$ into $K$ intervals $\{\Delta_1,..,\Delta_K\}$.
  \FOR{$\Delta_i \in \{ \Delta_1,..,\Delta_K$ \}}
      \item   Initialize $G_i$.
      \item  Create a vertex $v_j$ for each $A_j \in  \mathcal{A}$ with feature $X_j$.
      \item Create an edge $e_{ij}(t)$ between $v_i$ and $v_j$ if and only if $$S( P_i(\Delta_k), P_j(\Delta_k)) \geq \gamma.$$
      \item Assign $t = (k-\frac{1}{2})\delta$ and append $G_i$ to $\mathcal{G}$
     \ENDFOR
    \STATE return $\mathcal{G}$.
\end{algorithmic}
\end{algorithm} 

\section{Experiments}
This section presents an empirical evaluation of the proposed method, with a focus on two key research questions. {\bf Q1:} How does the performance of our approach compare with existing baselines? {\bf Q2:} Does incorporating multi-modal information enhance the solution?

\subsubsection{Data and Preprocessing}
The evaluation is carried out using publicly available financial data from Yahoo Finance and Naver Finance. The study primarily targeted three indices: the Korea Composite Stock Price Index (KOSPI), the Standard \& Poor's 500 Index (S\&P 500), and Heng Seng Index (HSI), thus offering a comparative study across different markets. We intentionally focus on the pre-pandemic period, specifically the daily closing price from 2015 to 2019, to circumvent the "black swan" effect induced by the COVID-19 pandemic. Information such as market capitalization, and sector information are used as elements in the feature vector of each entity. 

We divide data into training, validation and testing in a chronological manner. The data is split in a $60/20/20$ (\%) ratio. The first three years serve as the training data, the following year is used for validation, and the final year is used for testing. This arrangement ensures an evaluation setup that realistically mimics the chronological nature of financial data and allows for a rigorous evaluation

\subsubsection{Benchmark Methods}
Our approach is compared with four different methods which act as our experimental baselines. The first one utilizes a simple 2-layer MLP, depending solely on static entity features. The second method leverages Long Short-Term Memory (LSTM) \cite{sarmento2020enhancing}, which uses historical price data to identify correlations. The third method projects the time series into a time-frequency domain, utilizing a Convolutional Neural Network (CNN) for learning \cite{zhao2017convolutional}. We also include a traditional statistical method based on cointegration (COINT) \cite{sarmento2020enhancing}. All machine learning methods (including ours) are trained until convergence or a maximum of $50$ epochs, using the Adam optimizer with a learning rate of 0.01 and a weight decay of $0.0001$. 

\subsubsection{Evaluation Measures}
Our performance evaluation relies on two key measures: average precision (AP) and mean absolute percentage error (MAPE). AP is a standard measure for models predicting categorical outcomes, indicating the proportion of correctly identified correlated pairs in our case. Conversely, MAPE measures the precision in estimating the actual correlation value, giving us insight into how well the model predicts the correlation degree between identified pairs. 

\subsection{Results}

 {\bf Effectiveness.} The effectiveness of our proposed method, {\algoname}, is evaluated first. We selected a time interval of one month for both training and inference, enabling the capture of short-term volatility and long-term trends. The results of this experiment are presented in Table~\ref{tab:effectiveness} and the values are averaged from five independent trials. Bolded entries represent the best performance. $\uparrow$ indicates that higher values are better, and $\downarrow$ denotes that lower values are better. As Table~\ref{tab:effectiveness} shows, {\algoname} significantly outperforms the existing baselines in predicting future correlations among entities.

{\bf Ablation Study.} A ablation study is conducted to validate the design decisions of {\algoname}, examining the use of feature information and the structural information from the constructed graph. The impact of feature information is summarized in Table~\ref{tab:feature}. In {\algoname}-one-hot, feature vectors are replaced with a unique one-hot identifier for each entity. The results reveal a significant performance drop when feature information is removed. The impact of structural information is summarized in Table~\ref{tab:structure}. In {\algoname}-edgeless, the memory module is removed, and each entity's memory vector is used as input for the decoder. This eliminates the concept of temporal neighbourhood and hence the structural information. As Table~\ref{tab:structure} illustrates, removing structural information also leads to a performance reduction in {\algoname}.

\begin{table}
\vspace{-6mm}
\caption{Performance comparison of {\algoname} with baselines. }
\resizebox{0.5\textwidth}{!}{
\begin{tabular}{|c|c|c|c|c|c|c|}
\hline
     &  \multicolumn{2}{c|}{\textbf{KOSPI}}      &  \multicolumn{2}{c|}{\textbf{S\&P 500}} &  \multicolumn{2}{c|}{\textbf{HSI}} \\ 
\hline
     &   AP(\%) $\uparrow$   &  MAPE $\downarrow$  &  AP(\%) $\uparrow$   &  MAPE $\downarrow$ &  AP(\%) $\uparrow$   &  MAPE $\downarrow$\\
\hline
  COINT         &  55.8 ± 0.5  & 43.6 ± 1.2 & 56.6 ± 0.9 & 40.2 ± 1.3 & 51.4 ± 0.7 & 32.8 ± 1.2\\ 
MLP         &  48.2 ± 0.4  & 45.2 ± 1.6 & 46.8 ± 0.3 & 43.2 ± 1.4 & 44.2 ± 0.4 & 34.4 ± 1.3\\ 
  CNN         &  62.7 ± 0.4  & 32.8 ± 1.3 & 64.8 ± 0.3 & 37.2 ± 1.5 & 64.2 ± 0.8 & 25.8 ± 1.4\\ 
 LSTM         &  61.4 ± 0.3  & 30.6 ± 1.5 & 65.6 ± 0.5 & 34.8 ± 1.3 & 61.5 ± 0.5 & 23.3 ± 1.2\\ 
  {\algoname} (ours)         &  {\bf 72.8 ± 0.4}  & {\bf 24.2 ± 1.0} & {\bf 74.2 ± 0.4} & {\bf 27.8 ± 1.3} & {\bf 69.8 ± 0.7} & {\bf 16.8 ± 1.4}\\ 
\hline
\end{tabular}
}  
\label{tab:effectiveness}
\end{table}

\begin{table}
\vspace{-3mm}
\caption{Performance of {\algoname} w./w.o feature information}
\resizebox{0.5\textwidth}{!}{
\begin{tabular}{|c|c|c|c|c|c|c|}
\hline
     &  \multicolumn{2}{c|}{\textbf{KOSPI}}      &  \multicolumn{2}{c|}{\textbf{S\&P 500}} &  \multicolumn{2}{c|}{\textbf{HSI}} \\ 
\hline
     &   AP(\%) $\uparrow$   &  MAPE $\downarrow$  &  AP(\%) $\uparrow$   &  MAPE $\downarrow$ &  AP(\%) $\uparrow$   &  MAPE $\downarrow$\\
\hline
  {\algoname}-one-hot         &  63.2 ± 0.9  & 32.2 ± 1.2 & 64.3 ± 0.2 & 34.6 ± 1.5 & 60.8 ± 0.5 & 21.6 ± 1.9 \\ 
    {\algoname}         &  {\bf 72.8 ± 0.4}  & {\bf 24.2 ± 1.0} & {\bf 74.2 ± 0.4} & {\bf 27.8 ± 1.3} & {\bf 69.8 ± 0.7} & {\bf 16.8 ± 1.4}\\ 
\hline
\end{tabular}
}  
\label{tab:feature}
\end{table}

\begin{table}[!t]
\vspace{-3mm}
\caption{Performance of {\algoname} w./w.o structural information}
\resizebox{0.5\textwidth}{!}{
\begin{tabular}{|c|c|c|c|c|c|c|}
\hline
     &  \multicolumn{2}{c|}{\textbf{KOSPI}}      &  \multicolumn{2}{c|}{\textbf{S\&P 500}} &  \multicolumn{2}{c|}{\textbf{HSI}} \\ 
\hline
     &   AP(\%) $\uparrow$   &  MAPE $\downarrow$  &  AP(\%) $\uparrow$   &  MAPE $\downarrow$ &  AP(\%) $\uparrow$   &  MAPE $\downarrow$\\
\hline
  {\algoname}-edgeless         &  57.2 ± 1.1  & 36.2 ± 2.4 & 58.0 ± 0.9 & 38.2 ± 2.1 & 53.6 ± 0.8 & 26.2 ± 1.4 \\ 
  {\algoname}          &  {\bf 72.8 ± 0.4}  & {\bf 24.2 ± 1.0} & {\bf 74.2 ± 0.4} & {\bf 27.8 ± 1.3} & {\bf 69.8 ± 0.7} & {\bf 16.8 ± 1.4}\\ 
\hline
\end{tabular}
}  
\label{tab:structure}
\vspace{-2mm}
\end{table}
\section{CONCLUSION}
In this paper, we have embarked on an exploration of the application of deep learning in the realm of pair trading, a well-regarded quantitative investment strategy. This journey has led to the creation of a unique approach, {\algoname}, explicitly designed to amalgamate descriptive and time series data, thereby optimizing the process of discerning temporal correlations in pair trading. Our empirical evidence shows that {\algoname} is highly effective in automatically identifying correlated pairs, surpassing the performance of traditional baselines that rely exclusively on either descriptive or time series data.

Moreover, the novel integration of multi-modal information in our approach extends beyond the scope of this study and pair trading. Its potential impact is significant for other quantitative finance-related problems, and we look forward to seeing how this innovation could transform these areas in the future.

{\bf Acknowledgements.} This study was funded by the Yangtze River Delta Joint Research and Innovation Community (Grant No.2022CSJGG1200), National Natural Science Foundation of China 
(Grant No. 42302326) and the Central University Basic Research Business Special Fund Assistance (No. JZ2023HGTA0178).

\bibliographystyle{IEEEbib}
\bibliography{strings,refs}

\end{document}